# A Comparison Between Data Mining Prediction Algorithms for Fault Detection
# (Case study: Ahanpishegan co.)

Golriz Amooee[1*], Behrouz Minaei-Bidgoli[2], Malihe Bagheri-Dehnavi[3]

[1] Department of Information Technology, University of Qom
P.O. Box 3719676333, No.52, 24th avenue, 30 metri Keyvanfar, Qom, Iran

*Corresponding author

[2] Department of Computer Engineering, Iran University of Science and Technology
Tehran, Iran

[3] Department of Information Technology, University of Qom
Qom, Iran

**Abstract**

In the current competitive world, industrial companies seek to manufacture products of higher quality which can be achieved by increasing reliability, maintainability and thus the availability of products. On the other hand, improvement in products lifecycle is necessary for achieving high reliability. Typically, maintenance activities are aimed to reduce failures of industrial machinery and minimize the consequences of such failures. So the industrial companies try to improve their efficiency by using different fault detection techniques. One strategy is to process and analyze previous generated data to predict future failures. The purpose of this paper is to detect wasted parts using different data mining algorithms and compare the accuracy of these algorithms. A combination of thermal and physical characteristics has been used and the algorithms were implemented on Ahanpishegan's current data to estimate the availability of its produced parts.

**Keywords:** *Data Mining, Fault Detection, Availability, Prediction Algorithms.*

## 1. Introduction

In today's competitive world, improving reliability, maintainability and thus availability of industrial products becomes a challenging task for many companies. Reports indicated that performance and availability largely depends on reliability and maintainability. [14] There are many solutions to improve the maintenance of complex systems. One solution is corrective actions, including the required repair and maintenance activities after the occurrence of failures and downtimes. Because of the high costs, risks and time consumptions, this method is not appropriate. Another solution is to use systematic and planned repair methods. Although this method prevents from serious failures, but it could be very expensive. [13]

Given the issues above, to reduce costs and increase availability more effectively, it is better to predict errors before occurrence using data analysis. In industrial companies, since the generated data volume is growing there is a major need to process data in real time. New technologies, in term of both software and hardware, provide data collection from different resources, even when production rate is really high. In the company discussed in this paper, several sensors were installed on the devices to collect and review information during the operation. The information includes temperature, pressure and speed. Also a data base management system (DBMS) is used to control and managed data stored in the data base. [1]

Current methods of data analysis which work based on reviewing the appearance or annual statistical graphs, has many limitations to predict the performance and availability of the produced parts. So today, analysts are attending to use superior patterns to increase availability. In recent years, data mining is considered as one of the common methods for processing and discovering the hidden patterns. Tools such as data warehouses, data mining, and etc, provides new field for production and industry. So that by using these tools, companies can achieve competitive advantages. Specifically, through data mining - extracting hidden information from large data bases - organizations are able to predict future behaviors, and can make decisions based on knowledge. [2] Using data mining techniques to detect errors and inefficiencies, largely increase the capacity of equipments productivity. [5]

In this paper, we used data mining tools to identify defective parts in Ahanpishegan manufactory. Such analysis results in providing high quality products, improving produced parts and thus increase availability. The outline of the paper is as follows:

First we described required concepts briefly. Then we present an overview on the previous researches conducted to identify defective parts. We also frame the steps needed to create a model and apply data mining algorithms on Ahanpishegan's data. Finally, we provide some suggestions to improve the model for further studies.

## 2. Required concepts

### 2.1. Data mining

Data mining discovers hidden relationships in data, in fact it is part of a wider process called "knowledge discovery". Knowledge discovery describes the phases which should be done to ensure reaching meaningful results. Using data mining tools does not completely eliminate the need for knowing business, understanding the data, or familiarity with statistical methods. It also does not include clear patterns of knowledge. [2]

Data mining activities are divided into three categories:

1. **Discovery:** Includes the process of searching the database to find hidden patterns without a default preset.

2. **Predictive Modeling:** Includes the process of discovering patterns in databases and use them to predict the future.

3. **Forensic Analysis:** Includes the process of applying extracted patterns to fined unusual elements.

### 2.2. Prediction Algorithms

The purpose of a prediction algorithm is to forecast future values based on our present records. [3] Some common tools for prediction include: neural networks, regression, Support Vector Machine (SVM), and discriminant analysis. Recently, data mining techniques such as neural networks, fuzzy logic systems, genetic algorithms and rough set theory are used to predict control and failure detection tasks. [5] In this paper, the algorithms will forecast a probability for the given data situation. If the probability is equal to 1 it means the data (part) is normal, otherwise if the probability is equal to 0 the data (part) is considered non-conventional.

### 2.3. Failure

Failure means having faults, interruption or stop in a system which is shown as a deviation in one or more variable. The most common way to detect, predict and avoid failures is to collect and analyze the information produced during the time of operation and maintenance. [1] This detection, prediction, and avoidance of failures in early stages will increase the availability.

### 2.4. Fault Detection

Fault detection is defined as detecting abnormal process behaviors. [10] Fault detection techniques are divided into two categories: Model-based approaches [11] and Data-based approaches. [12] Since the construction of comparative models for real-time industrial processes is difficult, therefore model-based fault detection techniques are not as popular as data-based fault detection techniques. In recent years statistical tools such as PCA (Principal

Components Analysis), KPCA (Kernel Principal Component Analysis) and DPCA (Dynamic Principal Components Analysis) are widely developed.

## 3. Related works

Only a few articles exist in the field of identifying defective parts with the help of data mining tools. In an article Dr Shabestari review various types of defects in casting aluminum parts in the Aluminum Research Center of Iran. He concluded that a proper understanding of the characteristics of defective parts is a necessity for suppliers. He also concludes that the best way to solve this problem is to make a board of defects along with an example of defective parts and label each component with the name of corresponding fault. Mr. Ghandehari et al. used steel grain size to detect defective parts in terms of mechanical properties. They used non-destructive method of abysmal flows Instead of destructive methods of metallographic which was time consuming and costly.

Mr. Alzghoul et al. [1] used Data Stream Management Systems (DSMS) and data stream mining to analyze industrial data with the aim of improving product availability. They used three classification algorithms to investigate the performance and products availability of each algorithm. Algorithms which were used are: Grid-based Classifier, Polygon-based Classifier and One Class Support Vector Machine (OC-SVM). They concluded that OC-SVM accuracy (98%) is better than the two other algorithms.

In addition to this paper, other researches also used data stream mining for machine monitoring and reliability analysis [7], online failure prediction [8] and tool condition monitoring [9].

In other researches, Sankar et al. [10] used non-linear distance metric measured for OC-SVM to detect failures, Or Rabatel *et al*. [13] review monitoring sensors' data to find abnormalities and increase maintainability. Mr. Huang et al. [14] used statistical T2 and PCA methods to detect different failures in thermal power plants.

## 4. Research method

In this section we describe the approach which was used to predict defective parts. This article aims to identify defective parts by using multiple prediction algorithms and help the firm to maximize its productivity and increase its reliability and availability. Different steps of our work is described in next sections.

### 4.1. Understanding the data

The research presented in this paper is carried out in collaboration with Ahanpishegan Co., a manufacturer in automotive industry and a producer of car aluminum parts for companies like Sapco, Part tire, etc. This factory produced many parts such as engine bracket, feed bracket, Rear and front side brackets, etc. Due to the wide range of parts, we only focus on engine brackets. The general shape of this part and its relevant sizes, from two different angles, are shown in figure 1 and 2.

### 4.2. Data Preprocessing

Preparation and data preprocessing are the most important and time consuming parts of data mining. In this step, the data must be converted to the acceptable format of each prediction algorithm. First we find remarkable points about features and proportion of defective part, through interviews with managers and employees. For example, rising temperature has a large impact on corruption, or the rate of defective parts in last months of summer is higher. Then we identify outliers, cleanse data and ignore the constant variables by analyzing the current and past records and multiple interviews with experts and staff. Then we specify important fields which should be used in our prediction algorithms (our next step) to identify defective parts. Selected field for prediction algorithms, outliers and null values are shown in figure 3.

### 4.3. Creating artificial abnormal data

At this point we need to enter some distorted and corrupted data to measure the influential variables on performance. Also entering this data has a significant role in comparing the accuracy of our prediction algorithms. In this article we entered 10% defective part (100 records of our 1000 records).

### 4.4. Applying various prediction algorithms

Different kinds of trees such as CHAID, C&R, and QUEST along with other prediction algorithms including neural networks, Bayesian, logistic regression, and SVM has been applied on our data. The results are presented in next session.

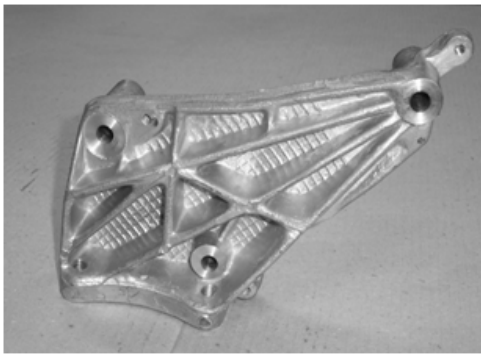
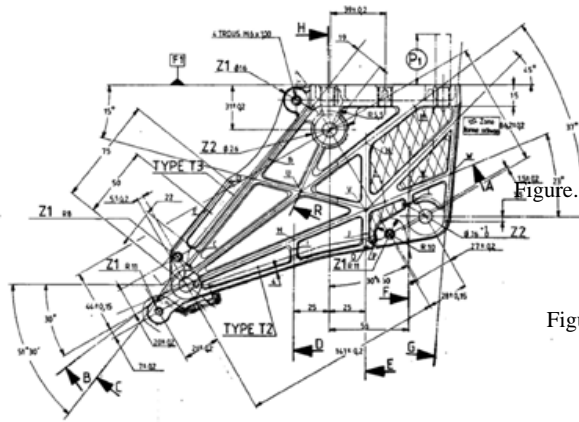

Figure.1 Engine bracket and its sizes f

Figure.2 Engine bracket and its siz

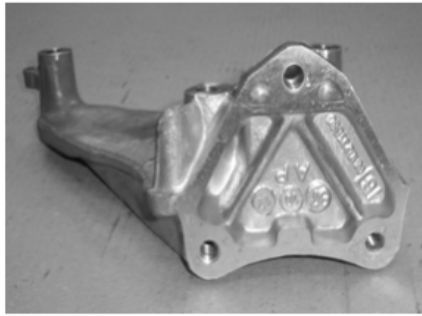
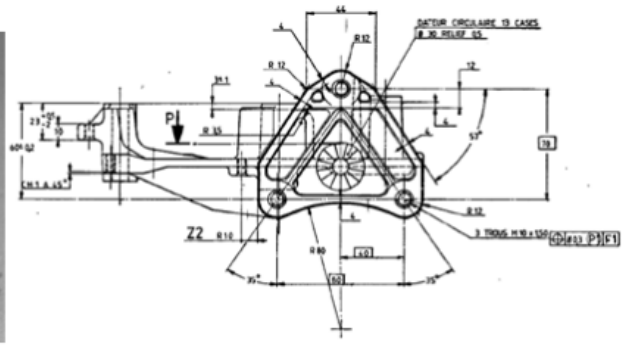

Figure.3 Final variables for pre

### 4.5. Investigating each algorithm's accuracy

The results of applying different algorithms are illustrated in figure 4. These algorithms are compared based on their accuracy and processing time. Due to data cleaning and removing noises and outliers, algorithms' accuracies are high.

Based on the result you can see that, SVM has the best processing time and also great overall accuracy. On the other hand algorithms which use trees to create their model needs more time and are sensitive to binary fields, but as you can see C&R and QUEST achieve the highest classification accuracy. A schema of a created C5 tree is shown in appendix A. Neural network is less accurate. Since our fields are numerical neural network has its own difficulties.

### 4.6. Rule generation using C5

After running different prediction algorithms and evaluating each of their accuracies, at this stage we aim to connect engine bracket features with known failures. So we run a C5 model on our data to generate prediction rules. An example of generated rules is as follows:

1) Mold temperature =< 325.500 and hardness <= 82 and distance between sensitive point and umbilical =< 23.95 then the part is normal.

2) Mold temperature > 325.500 and hardness > 82 and distance between sensitive point and umbilical =< 23.200 then the part is defective.

### 5. Conclusions and recommendations for future work

The purpose of this paper is to use data mining tools for identifying defective parts. The research presented in this paper has been carried out in collaboration with Ahanpishegan Co. First we find remarkable points about features and proportion of defective part, through interviews with managers and employees. Then by using an integrated database, identifying outliers, cleaning the data and ignoring the constant variables, we apply different prediction algorithms and compare the results. This paper aims to improve industrial product's reliability, maintainability and thus availability.

We recommend using data stream mining in future works to achieve quicker and more accurate results. Such researches can be performed in other companies such as food industry to measure products quality. Also future works can focus on frequency of failures, the cost of each failure and aim to minimize the consequence of such failures.

Figure.4 A comparison between algorithms

|   | Algorithm | Processing time (minute) | Accuracy (%) | Unused fields | Area Under Curve |
|---|---|---|---|---|---|
| 1 | CHAID | < 1 | 88 | 3 | 0.94 |
| 2 | Neural net | < 2 | 79 | 7 | 0.95 |
| 5 | C&R Tree | <5 | 92 | 6 | 0.92 |
| 6 | QUEST | <4 | 92 | 6 | 0.92 |
| 7 | Bayesian Network | <3 | 89 | 7 | 0.94 |
| 8 | Logistic regression | < 1 | 89 | 7 | 0.93 |
| 9 | SVM | < 1 | 90 | 7 | 0.92 |

**Appendix A**

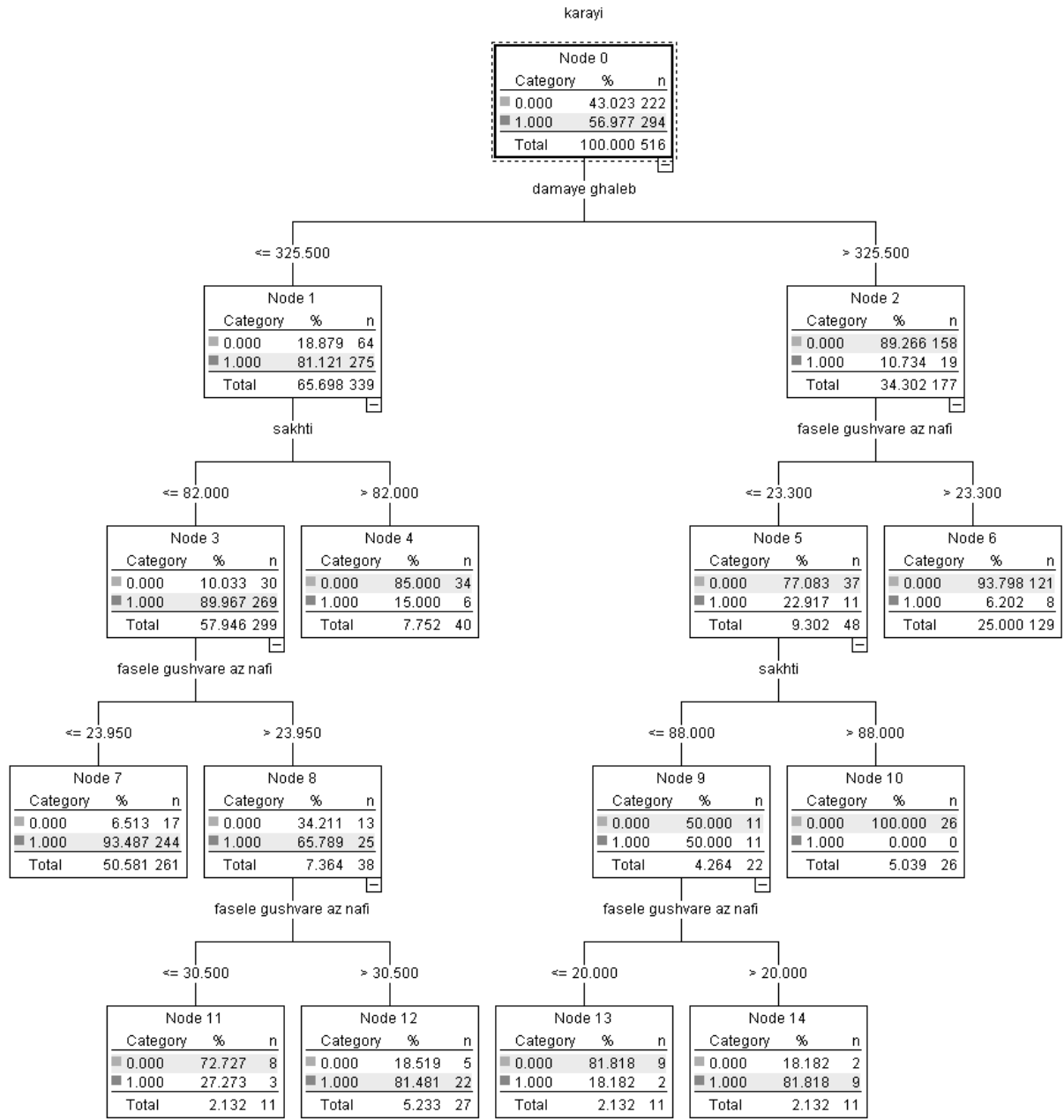

**Acknowledgments**

The authors would like to thank the referees for their valuable comments and suggestions, which greatly enhanced the clarity of this paper. We also would like to thank all the people that contributed with their expertise and work to the realisation of this research.

**Golriz Amooee** was born in Tehran, Iran in 1987. She received her B.S. degree with a first class Honors in Information Technology from Islamic Azad University, Parand Branch, Iran, in 2009 and currently she is a M.S. student in the Department of Information Technology at University of Qom, Iran. She specializes in the field of Customer Relationship Management (CRM), Information Security Management and ISO 27001.

**Dr Behrouz Minaei-Bidgoli** obtained his Ph.D. degree from Michigan State University, East Lansing, Michigan, USA, in the field of Data Mining and Web-Based Educational Systems in Computer Science and Engineering Department. He is working as an assistant professor in Computer Engineering Department of Iran University of Science & Technology, Tehran, Iran. He is also leading at a Data and Text Mining research group in Computer Research Center of Islamic Sciences, NOOR co. Qom, Iran, developing large scale NLP and Text Mining projects for Farsi and Arabic languages.

**Malihe Bagheri-Dehnavi** was born in Qom, Iran in 1988. She received her B.S. degree in and currently she is a M.S. student in the Department of Information Technology at University of Qom, Iran.